\pdfoutput=1
\documentclass[journal]{IEEEtran}
\usepackage{flafter}
\newcommand{\claimmaptab}{Table~S20}

\usepackage{amsmath,amssymb}
\usepackage{graphicx}
\usepackage{booktabs}
\usepackage{array}
\usepackage{url}
\usepackage{hyperref}
\usepackage{orcidlink}   %
\usepackage{tikz}
\usepackage{microtype}
\usepackage{algorithm}
\usepackage[noend]{algorithmic}
\usetikzlibrary{positioning,calc}
\definecolor{cLS}{HTML}{2166ac}
\definecolor{cCO}{HTML}{e08214}
\definecolor{cFI}{HTML}{b2182b}
\definecolor{cID}{HTML}{1b7837}
\definecolor{cIDn}{HTML}{5e3c99}   %
\definecolor{cPI}{HTML}{018571}   %
\newcommand{\aerr}{\alpha_{\mathrm{err}}}
\newcommand{\astar}{\alpha^{*}}
\newcommand{\IN}{I_{N}}

\newcommand{\OSFANON}{\url{https://osf.io/rjkwg/}}
\newcommand{\ZENODO}{will be archived at Zenodo under a versioned DOI on publication}

\begin{document}

\title{Wrong Operator or Blind Design?\\
A Post-Training Diagnostic for Physics-Informed\\ Coefficient Learning}

\author{Eric~Fock~\orcidlink{0000-0002-5017-3415}%
\thanks{E. Fock is with the PIMENT Laboratory, Université de La Réunion,
Le Tampon 97430, La Réunion, France (e-mail: eric.fock@univ-reunion.fr).}%
\thanks{ORCID: \url{https://orcid.org/0000-0002-5017-3415}}\thanks{This work has been submitted to the IEEE for possible publication. Copyright may be transferred without notice, after which this version may no longer be accessible.}}

\markboth{Preprint -- not peer reviewed}
{Wrong Operator or Blind Design?}

\maketitle

\begin{abstract}
Physics-informed and hybrid models integrate data and physics, and read learned coefficients as physical.
Yet a small prediction error guarantees neither that the postulated operator is right, nor that the design identifies the coefficient, nor that training reached its minimum.
On one parabolic benchmark where a sensor aggregates several diffusion rates and the truth is known, a correct operator recovers the coefficient to machine precision.
A wrong one leaves it $30\%$ off while the in-domain error is $1.36$ times the noise, and drifts $27\%$ across windows.
At $7\%$ noise the error falls below the noise; the coefficient stays $30\%$ off.
Without the truth, a post-training coefficient diagnostic questions the trained model with three classical statistics, without oracle; the first two need no refit.
Does it explain the data?
Can the design resolve the named physics?
Does the coefficient depend on the window?
The specification test keeps its pre-registered false-alarm ceiling and rejects every replicate of both wrong operators tested; a network frozen at its minimum returns the same verdicts.
Where the design is blind, a wrong operator confined to the unseen rate leaves the test mute; the design reading flags it beforehand.
A network stopped short of its minimum mimics a wrong operator; an analytic second fit separates the two.
The test has a notch against a missing reaction term, miscalibrates under correlated noise and loses power beyond $7\%$ noise.
Each warning names its remedy (operator, experiment or training); silence certifies nothing.
\end{abstract}


\IEEEpeerreviewmaketitle

\section{Introduction}\label{sec:intro}
\IEEEPARstart{D}{ata} and physics are now integrated ever more closely in learned models.
Physics-informed neural networks (PINNs) \cite{raissi2019,karniadakis2021} infer coefficients of partial differential equations (PDEs) from sparse, noisy data, with applications across science and engineering \cite{cuomo2022}.
So do hybrid models, which embed a first-principles structure and learn only its unknown components \cite{psichogios1992,vonstosch2014,schweidtmann2024,rackauckas2020}.
The workflow is well established \cite{jagtap2020,kapoor2024}.
One postulates the operator, parameterizes its coefficient and minimizes a loss of data misfit plus physics residual.
The coefficient is reported once the held-out error is small, and read as the scientific output of the model.

That reading rests on a silent hypothesis.
A small prediction error validates the fitted input-output map.
It guarantees neither that the postulated operator is right, nor that the design identifies the coefficient, nor that training reached its minimum.
The failure modes of training are well studied: ill-conditioning of the residual loss \cite{krishnapriyan2021}, gradient pathologies between loss terms \cite{wang2021}, spectral bias \cite{rahaman2019,wang2022ntk}, extrapolation failure \cite{fesser2023}.
All of them presuppose that the postulated operator generated the data.
This paper asks what a trained model can tell about its own coefficient when that is in doubt.

That premise fails in an ordinary way.
A sensor reads the sum of several diffusion processes with distinct rates: a heterogeneous medium probed at one point, several thermal paths behind one sensor, a mixture of decay rates.
A model lumps them into one effective rate, and nothing in the fit reports the aggregation.
On the exactly solvable problem studied here, the learned coefficient shows it, provided the truth is known.
Under a correct operator the scalar estimator recovers the diffusivity to machine precision ($2.2\times10^{-8}$, the single-precision rounding): prediction error and coefficient error go together.
Under a wrong operator they part.
At zero noise the in-domain error of a PINN degrades from $3.7\times10^{-8}$ to $5.3\times10^{-4}$, still an excellent fit, while the error on the diffusivity grows from $0.2\%$ to $30\%$.
This is not a failure of prediction but of attribution: the coefficient cannot be attributed to the system without further evidence.
With the correct structure restored, one rate per mode, they meet again: the rate error returns to $1.5\times10^{-6}$ and the out-of-domain error to $1.4\times10^{-14}$ (Section~\ref{sec:return}).

The $30\%$ plateau does not move with the noise (Section~\ref{sec:plateau}).
It has a name.
When the operator is wrong, the estimator converges not to the true coefficient but to a \emph{pseudo-true} value, the minimizer of the population loss under the wrong model \cite{huber1967,white1982}.
It follows the loss and the design, not the architecture.
It moves by $27\%$ across successive quarters of the window on the same field.
The recovered coefficient thus belongs to the pair (model, window), not to the medium.

A second difficulty leaves the same checks silent.
The design may leave a rate of a richer operator unconstrained.
The mode may have a vanishing amplitude, or have decayed before it is observed.
That rate is then not identifiable \cite{rothenberg1971,raue2009,wieland2021,ran2017}.
At zero amplitude more data would not improve an estimate of it, and the single fit of the postulated operator shows nothing.
Browning et al.~\cite{browning2025} describe this confusion for an estimated parameter.
Here it lies between a wrong operator and a design blind to a richer one.
The remedies are opposite: correct the operator, or change the experiment; a network short of its minimum calls for a third, more training.

The ladder above needs the truth, which the practitioner does not have.
A post-training coefficient diagnostic therefore puts three questions to the fitted model, with no oracle; the first two need no retraining of the network, the third refits once per window (Fig.~\ref{fig:classes}).
Does the fitted model explain the data?
An information-matrix test \cite{white1982} on the residuals answers: it compares the curvature of the loss with its squared score, both read in the physical parameter.
Can the design resolve the named physics?
A rank statistic derived from the Fisher information of the design answers, for a richer operator that the practitioner names and never fits.
Does the coefficient depend on the observation window?
A heterogeneity statistic on coefficients refitted over disjoint windows answers.
The three readings use only the learned coefficient, the residuals and the sensitivity $\partial_\alpha u$ of the postulated operator, in closed form, from a solver or by automatic differentiation.
They never use the weights, and therefore apply to any estimator that returns a coefficient.
Both wrong operators of the benchmark fire the first statistic in every replicate (Section~\ref{sec:notch} reports one it misses).
A wrong operator confined to a mode the design cannot reach leaves it mute, at $T_5<1$: the silence proves nothing, and the second reading says so before any fit of the richer operator (supplement, Section~S12).
The third shows that the coefficient of a wrong operator depends on \emph{when} one observes.

Sections~\ref{sec:bench} to~\ref{sec:networks} set the benchmark, pose the three questions, read them under a correct, a wrong and a restored operator, turn them into Algorithm~\ref{alg:proc} and read it on trained networks.
Section~\ref{sec:limits} reports three failures.
Power falls in a notch of the amplitude sweep and returns beyond it.
One reading is mute in some window regimes.
The neural estimator loses a pre-registered comparison.
Every decisional threshold was fixed before its run, with one declared exception and two cases where the commit chain does not attest the order (Section~\ref{sec:prov}).

The statistics are classical, with a lineage \cite{chesher1984,hall1987,golden2016,krauz2022} discussed in Section~\ref{sec:related}.
The contribution is not a new statistic.
It is a post-training coefficient diagnostic: classical identities read on the fitted model, in the physical parameter, and composed into a decision rule that separates a wrong operator, a blind design and an unreached minimum (Algorithm~\ref{alg:proc}).
The benchmark is exactly solvable on purpose.
Its aim is not performance on an application but a controlled test of these three readings, where the truth, the pseudo-true value and the sensitivities are computable.
Three results concern trained networks.
Their inductive bias does not move the estimate: perceptrons share the least-squares plateau, and a physics-informed network, whose loss is composite, settles $1.3\%$ from it (Section~\ref{sec:plateau}).
A network frozen at its minimum returns the verdicts of the analytic head, and only there (Section~\ref{sec:networks}).
\section{Related Work}\label{sec:related}
\textbf{How physics-informed estimators fail.}
The failure modes recalled in Section~\ref{sec:intro} improve with a better optimizer, more capacity or a reweighted loss, provided the target operator is true.
Ours does not: a one-parameter least-squares objective, minimised to its solver's tolerance, returns a coefficient wrong by a third (Section~\ref{sec:plateau}).

\textbf{Misspecification, and what has been done about it.}
The pseudo-true limit is recalled in Section~\ref{sec:intro}.
The calibration of imperfect models starts from the same object, a parameter that depends on the model, the loss and the design measure \cite{kennedy2001,tuo2015}; ignoring the discrepancy biases it \cite{brynjarsdottir2014}.
We do not claim that object.
We vary the design measure instead of holding it fixed: under a correct operator the pseudo-true value is the true one, whatever the design, so any dependence on the design becomes a test ($Q$).
In physics-informed learning the usual answer is to correct the model, by a learned discrepancy term \cite{zou2023} or by down-weighting the physics channel \cite{hua2024}; the corrector is run here as a baseline only (Section~\ref{sec:return}).
Misspecification has been tested from the observations alone as a composite hypothesis and compared with the information-matrix test \cite{krauz2022}; that work also finds a blind case \cite[Section~II-C]{krauz2022}.
In that linear Gaussian model the two information forms share one limit under both hypotheses; ours is nonlinear in the coefficient, with an estimated scale.
Section~\ref{sec:limits} reports the notch that our design does have.
Sequential, anytime-valid residual diagnostics detect structured patterns that a norm check accepts \cite{kazlauskaite2026}.
Like the tests above, they say nothing about whether the parameter was within reach.

\textbf{Identifiability, a separate axis.}
Identifiability analysis asks whether a parameter can be recovered at all \cite{rothenberg1971,raue2009,wieland2021}.
Ran and Hu survey it for this community \cite{ran2017}.
For linear-in-parameter parabolic operators, structural identifiability reduces to an auxiliary elliptic problem \cite{salmaniw2025}.
This literature diagnoses a flat likelihood whether or not the operator is right.
A recent data-only bound on what any discovery algorithm can recover \cite{gallo2026} rests on the invariant-measure moment matrix of a short trajectory.
That matrix is large, we note, whether the operator is right or wrong.
The misspecification literature, symmetrically, does not separate its plateau from a flat direction.
Closest to our question, Browning \emph{et al.}~\cite{browning2025} call the two failures confounded.
Their answer, a structural prior flexible enough to absorb the discrepancy, is chosen before the fit and lives with the confusion.
We report a verdict on the fitted model, after it.
The closest diagnostic of inverse-PINN coefficient error also reads two axes \cite{zhang2026}.
One is the resolution of a matched forward estimator (Fisher prediction).
The other is a signed residual displacement at the known truth.
Both are read on the frozen field and weights, and a two-by-two table routes them to the next experiment.
Our grid differs on three points.
Both statistics are read at the fitted rate, never at a truth.
The operator is actually wrong in two of our three configurations; theirs is matched and stressed only by a coarse solver.
And $T_5$ bears on the rate of a named alternative, not on the fitted parameter.

\textbf{Adjacent questions, kept apart.}
Out-of-distribution detection \cite{yang2024ood} asks whether a test input was drawn from the training law.
Here it was, and the postulated operator is what fails.
Amortized inference \cite{radev2022} detects misspecification as data outside the typical set of its simulations \cite{schmitt2024}.
Our test is on the fitted operator.
Benchmarks of learned PDE inverse solvers \cite{goel2026} vary the \emph{value} of the parameter under the operator fitted.
That is the other axis.
Architectures identifiable by construction \cite{yang2021,medina2024} make the parameterization identifiable.
The form may still be wrong (Section~\ref{sec:plateau}).

\section{Problem Setup, Designs and Estimators}\label{sec:bench}

\begin{table}[t]
\centering
\caption{The three generating configurations.}
\label{tab:scen}
\setlength{\tabcolsep}{3pt}
\resizebox{\columnwidth}{!}{%
\begin{tabular}{lccc}
\toprule
 & well-spec. & single-missp. & double-missp. \\
\midrule
modes & $[1,3,5]$ & $[1,3,5]$ & $[1,3,5,7]$ \\
amplitudes $c_n$ & $[1,0.3,0.1]$ & idem & $[1,0.3,0.1,0.05]$ \\
rates $\bar\alpha_n$ & $0.010$ (all) & $\{0.010,0.005,0.020\}$ & $\{0.010,0.006,0.018,0.014\}$ \\
config hash & \texttt{32cf4a5526cb} & \texttt{7935b218c150} & \texttt{0d7c972548f5} \\
postulated operator & \multicolumn{3}{c}{$\partial_t u=\alpha\,\partial_{xx}u$, one scalar $\alpha$} \\
\bottomrule
\end{tabular}}
\end{table}
\subsection{Governing equation, data and designs}\label{sec:designs}
We consider the one-dimensional heat equation on $[0,L]\times[0,T]$ with $L=1$ and homogeneous Dirichlet boundary conditions.
Its solution admits the modal expansion
\begin{equation}
u(x,t)=\sum_n c_n \sin\!\frac{n\pi x}{L}\,
\exp\!\Big(-\bar\alpha_n \big(\tfrac{n\pi}{L}\big)^2 t\Big).
\label{eq:modal}
\end{equation}
where $c_n$ and $\bar\alpha_n$ are the amplitude and the diffusion rate of mode $n$.
The initial condition is imposed, so the amplitudes are known; only \emph{cf2} (Section~\ref{sec:estim}) estimates them.
The identification target is the rate in \eqref{eq:modal}.
Throughout the paper we report the relative identification error $\aerr=|\hat\alpha-\alpha_0|/\alpha_0$ with respect to the nominal value $\alpha_0=0.01$.
The practitioner asks how far the reported rate is from the truth.
Observations consist of $u$ corrupted by i.i.d.\ Gaussian noise of \emph{absolute} standard deviation $\sigma$.
At $\sigma=0.02$ the field has an RMS amplitude of $0.69$, that is $2.9\%$ noise ($7\%$ at $\sigma=0.05$).
All models are trained on $t\in[0,1]$.
Prediction errors are computed against the noise-free field, which a practitioner does not observe.
In-domain and out-of-domain errors use grids of $50\times20$ and $100\times50$ points, the latter over $t\in[1,2]$.
Unless stated otherwise, every reported quantity is a median or a rate over $n=30$ independent replicates with data-generating seeds $42+13k$.
For a given seed all estimators see the same data, so all comparisons are paired.
Unless otherwise stated, intervals are $95\%$ bootstrap percentile intervals.

Table~S9 of the supplementary material specifies which quantity is computed on which design.
The \emph{noise sweep}, which produces the identification tables, uses $2000$ training samples at seven noise levels including $\sigma=0$.
We evaluate the diagnostic statistics on the \emph{detection design}.
It is a deterministic grid of $n_x=48$ equispaced interior nodes and $n_t=40$ midpoints of $[0,1]$, $N=1920$ points, at $\sigma=0.02$.
The information-matrix statistic is normalized by the sample variance of the residuals and cannot be calibrated at $\sigma=0$.

\subsection{Specification axis: three scenarios, one postulated operator}\label{sec:scen}
The specification axis is spanned by three content-hashed data-generating configurations, summarized in Table~\ref{tab:scen}.
Only the data change between them.
The operator postulated by every estimator, $\partial_t u=\alpha\,\partial_{xx}u$ with a single scalar rate, is the same in all three.
In the \emph{well-specified} configuration the field contains modes $[1,3,5]$ with a common rate.
A scalar operator is then exactly correct.
In the \emph{single-misspecified} configuration the same modes and amplitudes are generated with one rate per mode.
A scalar operator is then parametrically wrong, whereas a modal operator over $[1,3,5]$ remains correct.
In the \emph{double-misspecified} configuration a fourth mode with its own rate is added to the field.
A modal head's basis still spans $[1,3,5]$.

The double regime covers two effects.
For a modal head it is parametric, with one rate where several are present.
It is also structural, with a mode the basis cannot represent.
For a scalar head and for the PINN it is only parametric, since neither carries a basis that could be truncated.
The PINN receives the exact initial condition, fourth mode included, through one channel of its objective.
For this reason the two are not compared against each other along this axis.

\subsection{Identifiability axis and the resulting design grid}\label{sec:axes}
The second axis is a one-parameter family of \emph{correctly specified} designs.
We decrease the amplitude $c_5$ of the fifth mode from $0.1$ towards zero.
The rate $\alpha_5$ is then less and less constrained by the data.
At $c_5=0$ the solution no longer depends on it.
Three values are run, $c_5\in\{0.1,10^{-3},0\}$.
The design at $c_5=0.1$ is the well-specified design of the other axis.
Thus the two axes share one pivot, and all results of Section~\ref{sec:discrim} are read from it.
The diagnostic is read on three of the four combinations and on one instance of the fourth, misspecified on a blind design.
Its other instance is run and reported as a negative (Section~\ref{sec:discrim}).

\subsection{Estimators and their objectives}\label{sec:estim}
Estimators follow the two routes.
The hybrid route solves the postulated form and learns only the coefficient; the PINN route learns the field under the residual of the equation.
Seven estimators take the first route and are trained on every configuration of the specification axis.
\emph{Hyb\_scalar} is the $\chi$-architecture of Psichogios and Ungar \cite{psichogios1992}, in the form of \cite{hybmodel2004}.
It keeps the postulated form intact and its single rate is a bare trainable parameter.
The analytic sensitivity $\chi=\partial_\alpha u$ is injected into backpropagation.
\emph{MLP\_scalar} ($49$ parameters) and \emph{MLP\_scalar\_big} ($241$) replace the bare parameter by multilayer perceptrons.
Fed a constant input, they output the same scalar.
\emph{Hyb\_modal} carries one bare rate per mode over $[1,3,5]$.
It is therefore the correctly structured ansatz under single misspecification.
\emph{Hyb\_field} carries a spatially varying profile $\alpha(x)$ with $49$ parameters.
Finally, \emph{cf} is a nonlinear least-squares curve fit of the scalar rate that involves no network.
\emph{cf2} additionally frees the amplitudes $c_n$.
All of these estimators minimize the data misfit
\begin{equation}
\hat\alpha=\arg\min_{\alpha}\sum_i\big(u(x_i,t_i;\alpha)-y_i\big)^2 .
\label{eq:nls}
\end{equation}

The \emph{PINN} follows Raissi \emph{et al.}
\cite{raissi2019}.
It is a fully connected trunk $(x,t)\mapsto u$ with architecture $[2,64,64,64,1]$ and tanh activations.
Its $8578$ parameters include the rate, parameterized as $\alpha=\exp(\log\alpha)$.
It is trained with Adam for $5000$ epochs, then L-BFGS with $5000$ inner iterations.
Adam uses a tenfold larger learning rate on the physical parameter \cite{wang2021} and a plateau scheduler.
Its objective differs from \eqref{eq:nls}.
It comprises four channels, each a mean over its own set of points,
\begin{align}
\mathcal L(\theta,\alpha)={}&\lambda_{\mathrm{data}}\,\mathrm{MSE}\big(u_\theta,y\big)_{2000}\notag\\
&+\lambda_{\mathrm{pde}}\,\mathrm{MSE}\big(\partial_t u_\theta-\alpha\,\partial_{xx}u_\theta,\,0\big)_{2400}\notag\\
&+\lambda_{\mathrm{bc}}\big[\mathrm{MSE}(u_\theta(0,\cdot),0)_{200}+\mathrm{MSE}(u_\theta(L,\cdot),0)_{200}\big]\notag\\
&+\lambda_{\mathrm{ic}}\,\mathrm{MSE}\big(u_\theta(\cdot,0),u_0\big)_{200},
\label{eq:pinnloss}
\end{align}
with weights $\lambda=1/1/10/10$.
The boundary weight is applied once per boundary.
Since each channel of \eqref{eq:pinnloss} is a mean, a boundary or initial point weighs $5\times10^{-2}$ against $5\times10^{-4}$ for a data point.
These two channels are exact, noise-free and carry no information about $\alpha$.
Consequently, two channels estimate $\alpha$ while the two others constrain the shape of the network.
The minimizer of the composite objective is not characterized here.
\emph{PINN\_small} is the same estimator with a $[2,5,5,1]$ trunk of $52$ parameters.
\emph{PINN+corrector} adds the learned discrepancy term of Zou \emph{et al.}
\cite{zou2023} to the residual channel of the reference network.
It serves as a baseline in Section~\ref{sec:return}.
Fig.~\ref{fig:classes} groups the estimators according to the objective they minimize, which will turn out to be the relevant axis.

\begin{figure}[t]\centering
\resizebox{\columnwidth}{!}{%
\begin{tikzpicture}[font=\small, node distance=2.4mm and 3mm,
  box/.style={draw, rounded corners=1pt, inner sep=2.5pt, align=center, minimum height=6mm},
  lab/.style={font=\small\bfseries, anchor=west, inner sep=0pt},
  rd/.style={box, text width=27mm, minimum height=12mm},
  vd/.style={box, text width=27mm, font=\footnotesize, minimum height=8mm, fill=white},
  el/.style={font=\footnotesize, inner sep=1pt}]
\node[box, text width=16mm] (Pp) at (9mm,28.5mm) {postulated\\physics};
\node[box, text width=16mm] (Dd) at (9mm,20.5mm) {data\\(design)};
\node[box, text width=14mm] (Ll) at (37mm,24.5mm) {training};
\node[box, fill=black!6, draw=black!45, text width=19mm] (Gg) at (65mm,24.5mm) {\textbf{diagnostic}\\$\IN$, $T_5$, $Q$};
\node[box, text width=19mm] (Ii) at (103mm,24.5mm) {interpretation};
\draw[->] (Pp.east) -- (Ll.west); \draw[->] (Dd.east) -- (Ll.west);
\draw[->, thick] (Ll) -- (Gg); \draw[->, thick] (Gg) -- (Ii);
\node[el, above] at ($(Gg.east)!0.5!(Ii.west)$) {none flagged};
\draw[->, dashed, black!45] (Ll.north) -- (Ll.north |- 0,34mm) -- (Ii.north |- 0,34mm) -- (Ii.north);
\node[el, above, text=black!60] at (66mm,34mm) {usual shortcut: small error $\Rightarrow$ interpret};
\node[text=cFI, font=\bfseries] at (95mm,34mm) {$\times$};
\draw[->, black!65] ([xshift=-4mm]Gg.south) |- (37mm,16mm) -- (Ll.south);
\node[el, above, text=black!70] at (51mm,16mm) {retrain (gate)};
\draw[->, cIDn] (Gg.south) |- (9mm,11.5mm) -- (Dd.south);
\node[el, below, text=cIDn] at (50mm,11.5mm) {redesign the experiment ($T_5$)};
\draw[->, cFI] ([xshift=4mm]Gg.south) |- (-2mm,6mm) |- (Pp.west);
\node[el, above, text=cFI] at (18mm,6mm) {revise the operator ($\IN$, $Q$)};
\node[lab] (tr) at (0,0) {Training: any estimator, one fit};
\node[lab, text=cLS] (t1) at (0,-6mm) {(a) least-squares class, the hybrid route};
\node[box, anchor=north west] (xt1) at (0,-9mm) {$(x,t)$};
\node[box, fill=cLS!10, draw=cLS!60, right=of xt1] (sol) {postulated form $u(x,t;\alpha)$,\\closed form or solver};
\node[box, right=of sol] (u1) {$\hat u$};
\node[box, right=of u1] (o1) {$\sum_i(\hat u_i-y_i)^2$};
\node[box, below=of sol] (al) {$\alpha$: bare parameter, MLP output,\\per-mode rates, or a profile $\alpha(x)$};
\draw[->] (xt1)--(sol); \draw[->] (sol)--(u1); \draw[->] (u1)--(o1); \draw[->] (al)--(sol);
\node[lab, text=cPI!80!black] (t2) at (0,-31mm) {(b) composite class, the PINN route};
\node[box, anchor=north west] (xt2) at (0,-34mm) {$(x,t)$};
\node[box, fill=cPI!12, draw=cPI!65, right=of xt2] (net) {network $u_\theta$\\(trunk)};
\node[box, right=of net] (u2) {$\hat u$};
\node[box, right=of u2] (o2) {data $+$ residual($\alpha$)\\$+$ boundary $+$ initial};
\node[box, below=of net] (al2) {$\alpha$: one trainable scalar,\\entering the residual only};
\draw[->] (xt2)--(net); \draw[->] (net)--(u2); \draw[->] (u2)--(o2); \draw[->] (al2.east) -| (o2.south);
\node[lab] (at) at (0,-58mm) {Post-training diagnostic: no oracle; $\IN$ and $T_5$ on one fit};
\node[box, fill=black!6, draw=black!45, text width=95mm, anchor=north west] (in) at (0,-61mm)
  {learned coefficient $\hat\alpha$ (a network frozen at its minimum), residuals $r_i$,\\sensitivities $\partial_\alpha u$ and $\partial^2_\alpha u$ of the postulated form};
\coordinate (bus) at ([xshift=-4mm]in.north east);
\draw[thick] (o1.east) -| (bus); \draw[thick] (o2.east) -| (bus);
\draw[->, thick] (bus |- o2.east) -- (bus) node[midway, right, el] {$\hat\alpha$};
\node[rd, fill=cFI!8, draw=cFI!60, anchor=north west, minimum height=14mm] (IN) at ([yshift=-5mm]in.south west)
  {\textbf{Does the model}\\\textbf{explain the data?}\\\textcolor{cFI}{\textbf{$\IN$}}: specification};
\node[rd, fill=cIDn!8, draw=cIDn!60, right=7mm of IN, minimum height=14mm] (T5)
  {\textbf{Can the design}\\\textbf{resolve the physics?}\\\textcolor{cIDn}{\textbf{$T_5$}}: identifiability};
\node[rd, fill=cCO!10, draw=cCO!65, right=7mm of T5, minimum height=14mm] (Q)
  {\textbf{Does $\hat\alpha$ depend}\\\textbf{on the window?}\\\textbf{$Q$}: window invariance};
\draw[->] (in.south -| IN.north) -- (IN.north);
\draw[->] (IN) -- (T5) node[midway, above, el] {mute};
\draw[->] (T5) -- (Q) node[midway, above, el] {$\ge1$};
\node[vd, below=5mm of IN] (v1) {wrong operator\\$\to$ change the model};
\node[vd, below=9mm of v1] (v1b) {minimum not reached\\$\to$ retrain};
\node[vd, below=5mm of T5] (v2) {blind design $\to$ change\\the experiment; stop};
\node[vd, below=5mm of Q] (v3) {window-dependent\\estimand $\to$ the model};
\node[vd, below=2.5mm of v3] (v4) {none flagged\\on this design};
\draw[->] (IN) -- (v1) node[midway, right, el] {fires};
\draw[->] (v1) -- (v1b) node[midway, right, el, align=left] {optimization gate:\\refitted head mute};
\draw[->] (T5) -- (v2) node[midway, right, el] {$<1$};
\draw[->] (Q) -- (v3) node[midway, right, el] {fires};
\draw[->] (Q.east) -- ++(2mm,0) |- (v4.east) node[pos=0.25, right, el] {mute};
\end{tikzpicture}}
\caption{Training, then the post-training coefficient diagnostic, between learning and physical interpretation. Top: the workflow; the diagnostic sends each warning back to the operator, the experiment or the training, and its silence lets interpretation proceed on this design without certifying the model. Middle: the estimators, grouped by the objective they minimize. (a) Least-squares class, the hybrid route (\emph{Hyb\_scalar}, \emph{MLP\_scalar}, \emph{MLP\_scalar\_big}, \emph{Hyb\_modal}, \emph{Hyb\_field}, \emph{cf}, \emph{cf2}). (b) Composite class, the PINN route (\emph{PINN}, \emph{PINN\_small}, \emph{PINN+corrector}). Bottom: three questions put to the fitted model ($\IN$ of \eqref{eq:instat}, $T_5$ of \eqref{eq:rank}, $Q$ of Section~\ref{sec:qdef}), read left to right as in Algorithm~\ref{alg:proc}.}
\label{fig:classes}\label{fig:workflow}
\end{figure}

The PINN and the curve fit start at $\alpha=0.5$, the $\chi$-heads at $0.01$; under misspecification \emph{cf} and \emph{Hyb\_scalar} agree to six digits ($0.297041$, $0.222837$), so the start matters only in the well-specified column (Section~\ref{sec:cert}).

A textual description can drift from the code.
We therefore assert the architecture of each estimator (layer types, parameter count) at run time and in the test suite.

\section{Three Axes of a Post-Training Coefficient Diagnostic}\label{sec:instr}
This section poses the three questions of Fig.~\ref{fig:workflow}, each with its statistic, then the two checks practitioners already read.
The three axes answer three different failures: a wrong operator, a blind design and a window-dependent estimand.
On a network, the optimization state is a gate before them, not a fourth axis.
A network user already has the inputs: the learned coefficient and the predictions of the frozen network at the measurement points.
Automatic differentiation supplies the sensitivity of the postulated form when no closed form exists.
Algorithm~\ref{alg:proc} (Section~\ref{sec:discrim}) states the reading rule.

\subsection{Specification: does the model explain the data?}\label{sec:spec}
Let $r_i=y_i-u(x_i,t_i;\hat\alpha)$ be the residual at the $i$-th of the $N$ fitted points, and $\hat\sigma^2$ its sample variance.
The sensitivity is $\chi_i=\partial_\alpha u(x_i,t_i;\hat\alpha)$.
Under Gaussian errors the per-point score and Hessian contributions read
\begin{equation}
s_i=\frac{r_i\chi_i}{\hat\sigma^2},\quad
h_i=\frac{-\chi_i^{2}+r_i\,\partial^2_\alpha u(x_i,t_i;\hat\alpha)}{\hat\sigma^2},\quad
d_i=h_i+s_i^{2}. \label{eq:imparts}
\end{equation}
Summed over the points, $h_i$ and $s_i^2$ of \eqref{eq:imparts} give the Hessian of the loss and its empirical Fisher, two objects familiar from network optimization.
At a minimum they agree only if the model is correctly specified \cite[Sec.~4.2]{kunstner2019}.
This is White's information-matrix equality \cite{white1982}: $\mathbb E[d_i]=0$ when the operator is correct.
A structured residual breaks the cancellation.
Read at a minimum, their gap therefore tests the postulated physics.
The specification statistic is the studentized discrepancy
\begin{equation}
\IN=\frac{N\,\bar d^{\,2}}{\widehat V},\qquad
\bar d=\frac1N\sum_{i}d_i,\qquad
\widehat V=\frac1N\sum_{i}\bigl(d_i-\bar d\bigr)^{2}.
\label{eq:instat}
\end{equation}
All quantities in \eqref{eq:instat} are evaluated at the fitted pair $(\hat\alpha,\hat\sigma^2)$ of each replicate, so that no ground truth enters.
We retain the curvature term $r_i\partial^2_\alpha u$ rather than replacing $h_i$ by its Gauss--Newton approximation $-\chi_i^2$.
This term vanishes in expectation only under a correct operator.
Discarding it would silently change what the test measures under the alternative.
A parametric bootstrap at the fitted pair calibrates the null ($N_{\mathrm{boot}}=2000$, $95$th percentile, one threshold per regime).
The pre-registered ceiling on the well-specified rejection rate is $0.10$, with the Clopper--Pearson $95\%$ interval \cite{clopper1934}.
It bears on the \emph{observed} rate, not on the level of the test, which the bootstrap sets at a nominal $0.05$ (Section~S15).
The supplementary material gives the derivation, the sign convention, the conditions of the identity and the lineage of the statistic, the heteroscedasticity component of Hall's decomposition \cite{hall1987}.

\subsection{Identifiability: can the design resolve the named physics?}\label{sec:ident}
Let $H$ denote the $N\times3$ matrix of sensitivities $\partial u/\partial\alpha_j$ on the detection grid.
Let $R=H^{\top}H/\sigma^2$ be the total-sample information matrix, at the known noise level $\sigma$ of the design (the plug-in $\hat\sigma^2$ is used by $\IN$ only).
Partitioning the rate index into $\{5\}$ and the remaining rates $r$ and profiling the latter out leaves the Schur complement.
The identifiability statistic is
\begin{equation}
T_5=\alpha_5^{2}I_{55},\qquad
I_{55}=R_{55}-R_{5r}R_{rr}^{-1}R_{r5},
\label{eq:rank}
\end{equation}
that is, the squared ratio of $\alpha_5$ to the standard error the design assigns to it.
Two operators are named.
The practitioner has fitted the postulated one; the design is examined against the alternative, here one rate per mode.
We evaluate $H$ and $\alpha_5$ at the rates postulated for the alternative.
They come from the only fit available, the single fit of the postulated operator: $\alpha_5=\hat\alpha$.
Its value is $0.0100$ under correct specification and $0.0070$ under the wrong operator.
$T_5$ is thus the information the design carries for the alternative's rate near the fitted operator, as in a score test.
It is a property of the design, of the postulated rate and of the amplitudes $c_n$, taken as known from the initial condition; the true rates never enter.
Nor does it see how far the truth lies from that neighbourhood (Section~\ref{sec:windows}).
$T_5$ does not use the residuals, so it does not depend on which specification statistic is read.
A cheaper whiteness statistic fills the same role on these designs (Section~\ref{sec:discrim}).
$H$ has a fifth column proportional to $c_5$, and $R_{rr}$ does not depend on it.
Thus $I_{55}$ is exactly quadratic in $c_5$ (supplementary material, Section~S12).
A single evaluation then determines the entire family.

$\IN$ is a test, with a bootstrap null and a size.
The rule $T_5<1$ is a threshold of principle, a signal-to-noise ratio of one, without a null distribution.
$T_5$ grows with $N$ and $1/\sigma$, characterizes a design and is an \emph{ordering}, not a hypothesis test.

\subsection{Window invariance: does the coefficient depend on the window?}\label{sec:qdef}
The third reading refits the postulated operator on $K$ disjoint windows of the design.
It is the only one that re-estimates: on a network it needs one training per window, so it is not a computation on a frozen network alone.
The windows are fixed before the drift is seen, disjoint, and informative enough to identify the rate.
A rejection reads a window-dependent estimand: a wrong operator, or a coefficient varying in time.
Under a correct operator the pseudo-true value is the true coefficient on every window, by an identity given in Section~\ref{sec:discrim}.
We read Cochran's heterogeneity statistic \cite{cochran1954} on the $K$ estimates, $Q=\sum_k(\hat\alpha_k-\bar\alpha)^2/\mathrm{SE}_k^2$ with Fisher weights.
Each $\mathrm{SE}_k$ is taken at the residual variance of its own window as in \eqref{eq:imparts}.
$Q$ is calibrated by a parametric bootstrap with refit.

\subsection{Precision floor and prediction error}\label{sec:cert}
Practitioners read two checks: the error on the recovered coefficient, which needs the truth, and the prediction error.
Before reading the diagnostic, we fix their floor and exclude the trivial causes of the effects reported below.
These are inadequate training, insufficient capacity and optimization artefacts.
Three statements of increasing strength serve this purpose.

First, every estimator fits its own training window.
Under correct specification at $\sigma=0$, the in-domain prediction error is at the machine floor for all least-squares heads.
It equals $3.7\times10^{-8}$ for the PINN (tabulated in the supplementary material).
The PINN loses four orders of magnitude outside its window ($3.8\times10^{-4}$).
Indeed, its prediction is produced by the network.
The $\chi$-estimators instead predict through the analytic form and extrapolate with the accuracy of $\hat\alpha$ itself.

Second, capacity alone can produce a large identification error.
In the same column of Table~\ref{tab:aerr}, the reference PINN recovers the rate to $1.9\times10^{-3}$.
This is consistent with the sub-percent recovery reported in \cite{raissi2019}.
The under-sized PINN yields $0.137$ with the physics exactly correct, the lower rung of a capacity ladder: taken alone, an error of this magnitude says nothing about the specification.
With a correct operator and a fixed design, the out-of-sample identification error is $0.004\%$ with the coefficient alone.
It rises to $1.5\%$ when a free state network is co-estimated, and to $7\%$ when a spatial field $\alpha(x)$ is added.
The cost of network freedom is monotone in the amount of unconstrained network, without any misspecification.
This ladder, measured on a held-out seed block, does not bound the plateau reported below.

Third, every scalar estimator lies within a factor of $1.7$ of the information bound (Section~S2).
No claim rests on the under-sized PINN.

\begin{table}[t]
\centering
\caption{Identification error $\aerr$ at zero noise (median, $n=30$), against the
nominal $\alpha_0$.}
\label{tab:aerr}
\small\setlength{\tabcolsep}{3pt}
\begin{tabular}{lccc}
\toprule
Estimator & well-spec. & single-missp. & double-missp. \\
\midrule
Hyb\_scalar$^{\dagger}$ & $2.2\times10^{-8}$ & $0.297$ & $0.223$ \\
cf (curve fit, $\alpha$) & $1.3\times10^{-7}$ & $0.297$ & $0.223$ \\
cf2 ($\alpha+c_n$) & $3.6\times10^{-8}$ & $0.365$ & $0.268$ \\
Hyb\_modal (correct) & $3.3\times10^{-6}$ & $1.5\times10^{-6}$ & $2.7\times10^{-3}$ \\
Hyb\_field & $8.8\times10^{-6}$ & $0.419$ & $0.477$ \\
PINN & $1.9\times10^{-3}$ & $0.303$ & $0.226$ \\
PINN\_small & $0.137$ & $0.368$ & $0.355$ \\
\bottomrule
\end{tabular}
\par\vspace{2pt}
{\footnotesize $^{\dagger}$Equals $|\mathrm{float32}(0.01)-0.01|/0.01$ exactly on all $30$
seeds. The head starts at the nominal, where the zero-noise loss and its gradient vanish,
so the optimizer does not move. It reports the start at single precision. The
well-specified floor is read on \emph{cf} and the PINN, which start far from the answer.\par}
\end{table}

The curve fit \emph{cf}, which has no network, lands on the plateau of its class (Table~\ref{tab:aerr}, Section~\ref{sec:plateau}).
Capacity alone costs $14\%$, the operator $30\%$, and the estimator without capacity incurs the latter.

Model-comparison tools do not address the question.
Cross-validation and information criteria \emph{compare} candidates, and a comparison has no null hypothesis.
It returns a winner whether or not any candidate is correct.
A nested test differs: once a richer operator is named (Section~\ref{sec:instr}), a score test of the postulated restriction has a null hypothesis.
Moreover, cross-validation is blind by construction.
The pseudo-true value is the minimizer of a risk integrated over the design measure.
Every fold is drawn from the same measure and hence shares the same pseudo-true value \cite{white1982}.
By contrast, $\IN$ interrogates a single fit for its internal consistency, with no competing model.

\section{Correct Operator: the Coefficient at its Floor}\label{sec:zero}\label{sec:calib}
On designs of known status the coefficient is read first, then the instrument.
Under the correct operator the coefficient sits at its floor and the instrument at its zero; under the wrong one, on the plateau while the instrument fires; with the structure restored, both return.
Under a correct operator the target of identification is the parameter itself, and the estimators recover it to their precision floor: to machine precision for the scalar head ($2.2\times10^{-8}$), the under-sized PINN excepted (Table~\ref{tab:aerr}).

Under correct specification, on the detection design ($\sigma=0.02$, Section~\ref{sec:designs}), the median is $\IN=0.193$.
The rejection rate is $0.033$ against the bootstrap threshold $2.41$, below the pre-registered ceiling of $0.10$ (Table~\ref{tab:white}).
This is the reference level against which all other readings are interpreted.
It is the zero of the \emph{statistic}, not of the estimator.

\begin{table}[t]
\centering
\caption{Information-matrix statistic on a single fit, $n=30$, at the detection design
($\sigma=0.02$, plug-in scale, per-seed $\hat\alpha$). Threshold: bootstrap $95$th
percentile per regime; $\chi^2(1)$ reference $3.84$. The last three rows report the same
reading when the deployed field is a trained network frozen before testing. Both
misspecified network cells were run under a separate pre-registration, in sample only.}
\label{tab:white}
\setlength{\tabcolsep}{4pt}
\begin{tabular}{lcccc}
\toprule
Regime & median $\IN$ & rejection (in) & (OOS) & threshold \\
\midrule
\multicolumn{5}{@{}l}{\emph{Least-squares surrogate} (analytic field)}\\
well-specified & $0.193$ & $0.033$ & $0.10$ & $2.41$ \\
single-missp. & $223.9$ & $1.00$ & $1.00$ & $2.66$ \\
double-missp. & $85.0$ & $1.00$ & $1.00$ & $2.59$ \\
\midrule
\multicolumn{5}{@{}l}{\emph{Trained network, frozen before testing}}\\
well-specified & $0.220$ & $0.067$ & $0.133$ & $2.41$ \\
single-missp. & $130.6$ & $1.00$ & -- & $2.67$ \\
double-missp. & $39.3$ & $1.00$ & -- & $2.58$ \\
\bottomrule
\end{tabular}
\par\vspace{2pt}
{\footnotesize All pre-registered claims met on the surrogate. On the network, power met
on both misspecified regimes, and specificity in sample. Out of sample, $0.133$ was not a
decision criterion for that cell. A pre-registered extension to $200$ seeds, containing these thirty, gives
$0.065$ ($[0.035,0.109]$). That is $4$ of the $30$ shared seeds and $9$ of the $170$ new
(Section~S3).
Both blocks share one convention
(Section~\ref{sec:zero}); thresholds agree to $0.05\%$ under
correct specification and $0.4\%$ under misspecification. Two computation paths agree
to $\le10^{-6}$.\par}
\end{table}

Out of sample the well-specified rejection rate equals $0.10$, at the ceiling rather than below it; a $200$-seed repetition narrows the interval without moving the boundary.
We report rather than repair the \emph{failed} pre-registered $\chi^2(1)$ calibration of the plug-in statistic; validity rests on the bootstrap (Sections~S13 and~S15).
That null was pre-registered after the $\chi^2(1)$ criterion failed, before its own first run.
The medians are properties of the design; only the crossing of the threshold is decisional.

On the same design the identifiability statistic equals $T_5=2159$, so the design resolves $\alpha_5$ by three orders of magnitude.
Both statistics thus flag nothing on the design where nothing is wrong.

Swept over six noise levels $\sigma\in\{0.005,\dots,0.2\}$ on the surrogate, the statistic rejects every single-misspecified replicate up to $\sigma=0.05$, that is $7\%$ of the RMS amplitude of the field.
It then falls to $0.30$ at $\sigma=0.1$ and to $0.10$ at $0.2$.
Under correct specification the false-rejection rate is at most $0.033$ at every level.
The statistic therefore fails by silence rather than by false alarm.
The bootstrap calibration also holds away from the noise level where it was set (sweep, double misspecification included, in the supplementary material).
\begin{table}[t]
\centering
\caption{Family invariance within the least-squares class, and the PINN's departure.
Single misspecification, $\sigma=0$, $n=30$, in sample. \emph{This cell has its own
training budget} (Table~S9 of the supplementary material). Its PINN value is not the same
run as the PINN column of Table~\ref{tab:aerr}.}
\label{tab:family}
\setlength{\tabcolsep}{3pt}
\begin{tabular}{lccc}
\toprule
Estimator & params & $\hat\alpha$ (median) & bootstrap CI$_{95}$ \\
\midrule
cf (curve fit) & $1$ & $\equiv$ Hyb\_scalar & -- \\
Hyb\_scalar (bare) & $1$ & $0.0070284$ & $[0.0070178,0.0070402]$ \\
MLP\_scalar & $49$ & $0.0070296$ & $[0.0070190,0.0070408]$ \\
MLP\_scalar\_big & $241$ & $0.0070296$ & $[0.0070190,0.0070408]$ \\
\midrule
PINN (composite) & $8578$ & $0.0069391$ & $[0.0069141,0.0069567]$ \\
\bottomrule
\end{tabular}
\par\vspace{2pt}
{\footnotesize Least-squares pseudo-true $\astar=0.007033$.\par}
\end{table}

\section{Misspecified Operator: the Plateau}\label{sec:plateau}
Under a single misspecified rate the scalar estimator converges to $\aerr=29.7\%$ at $\sigma=0$ and $29.8\%$ at $\sigma=0.02$.
We call it the pseudo-true coefficient plateau.
What matters is not its value but its invariance within one objective (Table~\ref{tab:family}): changing the architecture without changing the estimand does not restore physical validity.

The plateau is not a noise effect.
Over the six noise levels $\sigma>0$ the log--log slope of the out-of-domain MSE of the scalar estimator is $0.0049$, with interval $[0.0019,0.010]$.
Under correct specification $\aerr$ is linear in the noise level (slope $1.007$); misspecification drives both slopes to zero, and a noise-invariant bias separates the regimes (Section~S10, with the variance).

\begin{figure}[t]\centering
  \includegraphics[width=\columnwidth]{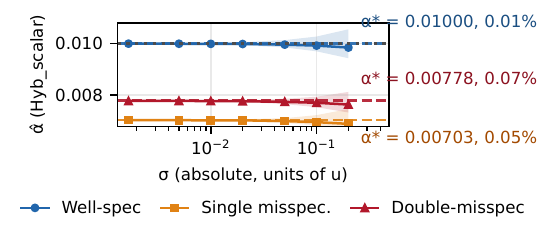}
  \caption{Empirical plateau versus White's pseudo-true $\astar$ (closed-form objective, no network). The median identification settles on $\astar$ to within $0.07\%$ across the three regimes.}
  \label{fig:white}
\end{figure}

The plateau is not a numerical coincidence either.
It coincides with the pseudo-true parameter $\astar$ of quasi-maximum-likelihood theory \cite{white1982}.
$\astar$ is the scalar minimiser of a closed-form objective, without any network: $0.01000$ / $0.00703$ / $0.00778$.
Empirical medians are $0.010000$ / $0.007030$ / $0.007772$ in the three regimes (Fig.~\ref{fig:white}).
The discrepancies are not a lack of convergence.
Under a correct operator $\astar=\alpha_0$ for every design (Section~\ref{sec:discrim}), and the $0.01\%$ is the tolerance of the scalar solver.
Under misspecification, the $0.05\%$ and $0.07\%$ reflect the gap between the continuous design measure and the $2000$-point training design.

Within the least-squares class, the plateau does not depend on the architecture.
A curve fit, a bare parameter and perceptrons of $49$ and $241$ parameters all minimize \eqref{eq:nls}.
Table~\ref{tab:family} reports their estimates and bootstrap intervals.
They converge to a single value with overlapping intervals.
Fed a constant input, the perceptrons establish that an over-parameterised minimiser of \eqref{eq:nls} reaches the same argmin.
They do not establish an invariance to capacity.
The one member with a genuine nuisance, the field head $\alpha(x)$, settles elsewhere ($42.0\%$ against $29.8\%$, Table~S2).
The PINN does not join this cluster.
Its objective \eqref{eq:pinnloss} defines another M-estimation problem, hence another pseudo-true value.
The PINN interval is disjoint from the cluster by $6.11\times10^{-5}$.
Only the disjointness is used, as a control.
This invariance holds within a loss class.
The departure of the PINN is what makes it informative rather than tautological.
Heads identifiable by construction \cite{yang2021,medina2024} were not run.
If they minimise \eqref{eq:nls}, they reach the same, wrong, $\astar$.

Across the ablation cascade (supplementary material) the plateau follows the \emph{ansatz}.
It is present for the scalar and field heads, absent for the modal head, and at its own level for the PINN.

On this design the specification statistic responds as expected.
$\IN$ rejects in every replicate on the detection design, both in sample and on the disjoint seed block (Table~\ref{tab:white}).
Its median is $223.9$ under single and $85.0$ under double misspecification, two orders of magnitude above the threshold.
On the single-misspecified design, at the postulated rate ($\hat\alpha=0.0070$), the identifiability statistic is $T_5=2378$.
The operator is wrong, but the design is not blind to the alternative.

For completeness we report what the usual accuracy check returns on this cell.
The in-domain prediction error of the misspecified \emph{MLP\_scalar} head is flat in $\sigma$.
Its RMSE is $0.0273$ from $\sigma=0.01$ to $0.20$.
At $\sigma=0.02$ it is $1.36$ times the observation noise.
A practitioner sees it only through the residual, where it adds to the noise in quadrature.
For $\sigma\ge0.05$ it lies \emph{below} the noise (RMSE$/\sigma=0.55$, $0.27$, $0.14$).
Yet the recovered coefficient is wrong by $30$ to $31\%$ throughout (Table~S8 of the supplementary material).
From $\sigma=0.05$ upward, therefore, no check based on predictive accuracy alone can flag the operator unless $\sigma$ is known.
The residual variance then exceeds $\sigma^2$ by $30\%$ at $\sigma=0.05$, or $6.6$ standard errors on the $1000$ points of Table~S8.
It exceeds it by only $7.5\%$ and $1.9\%$ at $0.1$ and $0.2$.
\begin{figure}[t]\centering
  \includegraphics[width=\columnwidth]{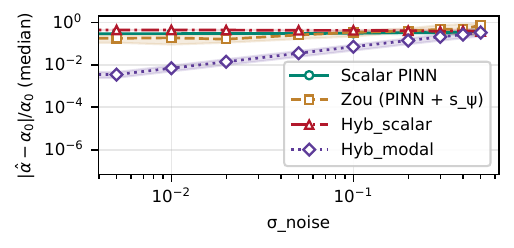}
  \caption{Identification error against noise. Scalar PINN, corrector baseline (PINN plus learned discrepancy \cite{zou2023}), scalar $\chi$-head and correctly structured modal head. Median over $n=30$ paired seeds. The modal head
  dominates the corrector up to $\sigma\approx0.3$ under Holm correction; beyond, the
  gap is no longer certified at $n=30$.}
  \label{fig:zou}
\end{figure}

\begin{table}[t]
\centering
\caption{Discrimination on one fit: two readings, two axes. $n=30$, detection
design, pre-registered thresholds reused (ceiling $0.10$; $T_5<1$, anteriority not attested,
Section~\ref{sec:prov}). The first row is
the pivot, bit-identical to the well-specified row of Table~\ref{tab:white}. Rows
descend along the identifiability axis; the last three rows are wrong-operator designs.
$T_5$ at the rate of the single fit ($0.0100$; $0.0070$ in the fourth row, where the true
rate $0.020$ would give $1237$).}
\label{tab:discrim}
\setlength{\tabcolsep}{4pt}
\begin{tabular}{lccc}
\toprule
Design & $\IN$ rejection & median $\IN$ & $T_5$ (rank) \\
\midrule
correct, $c_5=0.1$ (pivot) & $0.033$ & $0.193$ & $2159$ \\
correct, $c_5=10^{-3}$ & $0.067$ & $0.387$ & $0.216$ \\
correct, $c_5=0$ (non-identifiable) & $0.067$ & $0.384$ & $0.000$ \\
\midrule
wrong operator, $c_5=0.1$ & $1.00$ & $223.9$ & $2378$ \\
wrong op.\ (mode $5$), $c_5=10^{-3}$ & $0.067$ & $0.389$ & $0.216$ \\
wrong op.\ (modes $1,3$), $c_5=0$ & $0.633$ & $3.45$ & $0.000$ \\
\bottomrule
\end{tabular}
\end{table}

\section{Restored Operator: the Coefficient and the Zero Return}\label{sec:return}
This last calibration step is the restoration, a reversibility control.
On the single-misspecified data the modal head is a correct operator, so the instrument must return to its zero.
The plateau disappears once the correct structure is restored.
Given one rate per mode over $[1,3,5]$, the modal head recovers the per-mode rates out of sample.
They are $\{0.01000,\,0.00500,\,0.02000\}$, with errors below $10^{-4}\%$.
Its out-of-domain MSE returns to $1.4\times10^{-14}$ (supplementary material).
The return is also direct on the $\IN$ channel.
For the modal head, $\IN$ takes its matrix form ($p=3$, six indicators).
Under single misspecification it rejects in $2$ of $30$ seeds, and in $0$ of $30$ under correct specification.
It rejects in $30$ of $30$ when a fourth rate is present in the field but absent from the head.
On that last design the rejections come from the first quarter of the window only ($0.97$ on $[0,\tfrac14]$, then $0.00$, $0.03$, $0.00$).
Indeed, the missing mode has decayed by $t\approx0.3$.
For the data observed afterwards, the three-rate operator is adequate (Section~\ref{sec:limits}).
The identifiability statistic is not evaluated on this rung, since this step is a control along the specification axis.

An alternative to restoring the structure is to keep the wrong operator and add a learned discrepancy term \cite{brynjarsdottir2014,zou2023}.
We run this corrector as a baseline.
At $\sigma=0.02$ the correctly structured modal head reaches an error of $1.41\%$.
The corrector reaches $16.5\%$.
Over the sweep $\sigma\in\{0,\ldots,0.5\}$ the paired comparison is significant for $\sigma\le0.3$, not for $\sigma\ge0.4$ (Section~S15).
The advantage is not claimed beyond $\sigma=0.3$ (Fig.~\ref{fig:zou}).
One asymmetry works against the corrector.
From $\sigma=0.1$ onward it performs \emph{worse} than the uncorrected network, with median ratios rising from $1.10$ to $1.86$ at $\sigma=0.5$.
\begin{figure*}[t]\centering
  \includegraphics[width=\textwidth]{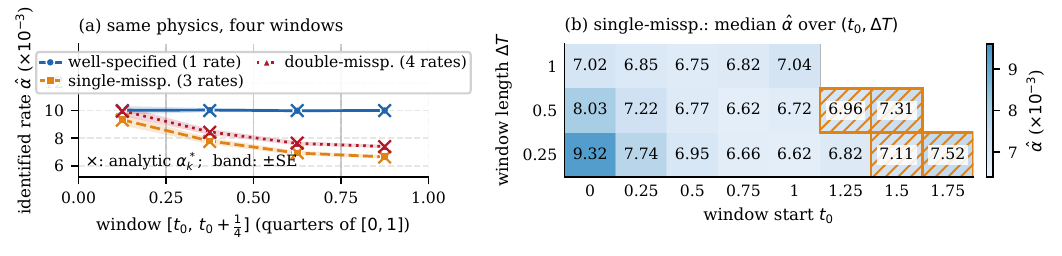}
\caption{The observation-window axis. (a) Same generating field, four disjoint quarters of $[0,1]$ (their union is the detection design, same noise). Median identified rate of the scalar operator over $n=30$ seeds (dots), $\pm$ one Fisher standard error (band). Analytic pseudo-true value per window (crosses, computed before the run). Invariant by identity under a correct operator, drifting by $27\%$ of $\alpha_0$ under aggregated rates. (b) Median identified rate over sliding windows $(t_0,\Delta T)$ on $[0,2]$ under single misspecification. The correct operator reads $10.00$ in every cell ($9.97$--$10.02$, same windows and seeds). Hatched cells: $T_5<1$ at the generating rates of the field, an oracle reading. The deployable reading at the postulated rate stays $\ge286$ everywhere (Section~\ref{sec:windows}).}
\label{fig:window}
\end{figure*}

\section{Three Causes: Wrong Operator or Blind Design}\label{sec:discrim}
\begin{algorithm}[!t]
\caption{Post-training coefficient diagnostic (output: revise the model, redesign the experiment, retrain the estimator, or none flagged on this design)}\label{alg:proc}
\begin{algorithmic}[1]
\footnotesize
\REQUIRE data on a design; the postulated operator, its sensitivity $\chi=\partial_\alpha u$
(closed form or solver); $\hat\alpha$ from any estimator, a network frozen at its minimum;
a richer alternative with a rate $\alpha_5$; i.i.d.\ noise
\STATE compute $\IN$ at $(\hat\alpha,\hat\sigma^2)$; threshold: parametric bootstrap
\IF{$\IN$ fires}
  \IF{network \AND $\IN$ of the analytic head refitted on the same data is mute}
    \STATE \emph{minimum not reached}: retrain, restart at line~1
  \ENDIF
  \STATE \emph{wrong operator}: change the model, restart at line~1
\ENDIF
\STATE compute $T_5=\alpha_5^2I_{55}$ at the postulated rate
\IF{$T_5<1$}
  \STATE \emph{design blind to $\alpha_5$}: the silence proves nothing; change the experiment; \textbf{stop}
\ENDIF
\STATE refit on $K$ disjoint windows; compute $Q$; threshold: bootstrap with refit
\IF{$Q$ fires}
  \IF{network \AND $Q$ of the analytic head on the same windows is mute}
    \STATE \emph{window-dependent optimisation}: retrain, restart at line~1
  \ENDIF
  \STATE \emph{window-dependent estimand}, a wrong operator: change the model, restart at line~1
\ENDIF
\STATE none flagged on this design
\end{algorithmic}
\end{algorithm}
Three causes can leave a learned coefficient wrong under a small error: a wrong operator, a design blind to the physics one names, and an optimization stopped short of its minimum.
This section separates the first two; Section~\ref{sec:networks} the third.
Algorithm~\ref{alg:proc} turns the three readings into a decision.
A rejection of $\IN$ reads a wrong operator, and the remedy is to change the model.
A silence is read against the design.
The diagnostic is deliberately asymmetric: no rejection is not evidence that the operator is correct, only that this design and these readings found nothing against it.
The causes are not exclusive; all three readings are reported.
If $T_5<1$, the design is blind to the alternative's rate and the remedy is the experiment.
Otherwise the window reading decides whether the coefficient depends on when one observes.

\textbf{The identifiability axis.}
$T_5$ separates an unresolvable rate ($c_5=0$, not identifiable \cite{rothenberg1971}) from a poorly resolved one ($c_5>0$).
The latter is identifiable but poorly estimable \cite{jacquez1985,raue2009}; cf.\ the resolution axis of \cite{zhang2026}.
Nobody estimates that rate; the design is examined for it.
Sections~\ref{sec:zero} to~\ref{sec:return} read the diagnostic along the specification axis with $c_5$ at its nominal value.
Here we read it along the identifiability axis with the operator held correct, moving away from the shared pivot design.
We show that the two statistics diverge.

Down the column of Table~\ref{tab:discrim}, $c_5$ decreases from $0.1$ to $10^{-3}$ to $0$.
The rank statistic falls from $2159$ to $0.216$ to exactly zero, that is by four orders of magnitude.
It crosses $T_5=1$ between the first and second rows and ends in a structural zero, a vanishing column of $H$.
Along this column the specification statistic does not move: $0.033$, $0.067$ and $0.067$.
All are below the ceiling (Clopper--Pearson $[0.008,0.221]$ for the last two).
The intermediate row is predicted rather than observed.
The quadratic law gives $2159\times(10^{-3}/0.1)^2=0.2159$ against the certified $0.2159$.
$T_5$ predicts the standard error of a fit of the richer operator, $\alpha_0/\sqrt{T_5}$.
Performed once to check (modal head, same data, pre-registered, $n=30$), that fit returns $0.021\,\alpha_0$ at $c_5=0.1$ for a predicted $0.022$, with a seed-to-seed scatter of $0.90$ of it.
At $c_5=10^{-3}$, $12$ of $30$ fits sit on a bound, and the standard error spans $0.4$ to $23\,\alpha_0$ (median $1.88$).
At $c_5=0$ the rate is undetermined (Table~S10 of the supplementary material).

Along the row (pivot to misspecified design), the specification statistic rises from a rejection rate of $0.033$ to $1.00$.
Its median goes from $0.193$ to $223.9$.
The rank statistic moves from $2159$ to $2378$, a factor of $1.1$ set by the postulated rate ($0.0100$ against $0.0070$).
Both values stay three orders of magnitude above the threshold.
Each statistic thus moves by at least three orders of magnitude along one axis and stays on the same side of its threshold along the other.
The fourth combination, a wrong operator on a design blind to the alternative, has two instances.
With the wrong rate on the unseen mode only, it was pre-registered after this table was frozen and run against those thresholds.
There $\IN$ is mute ($2$ of $30$) at $T_5=0.216$, the silence cell of Section~I (fifth row; supplement, Section~S12).
With the wrong rates on the visible modes and $c_5=0$, it is run: a pre-registered negative on power at $T_5=0$ (last row; Table~\ref{tab:negatives}).
In the $11$ mute seeds Algorithm~\ref{alg:proc} returns ``blind design'', which it is, never ``correct operator''.
The near-invariance of $T_5$ along the specification axis rests on three designs; it was not evaluated on the double-misspecified one.

The identifiability boundary follows from the quadratic law.
On the misspecified design, at the postulated rate $0.0070$, $T_5=1$ is reached at $c_5^{*}=0.1/\sqrt{2378}=2.05\times10^{-3}$.
At the true rate $0.020$ it would be $2.84\times10^{-3}$.
On the correct design it is reached at $2.15\times10^{-3}$.
The intermediate value $c_5=10^{-3}$ lies below both boundaries.

At $n=200$ the specificity bound closes in sample: on the non-identifiable design the rejection rate is $0.050$ ($10/200$), and out of sample $0.075$ ($15/200$).
The discrimination is therefore a \emph{bound} in sample and a \emph{direction} out of sample (Section~S9).

\textbf{The observation-window axis.}
The identifiability axis above changes the generating field through $c_5$.
A reader may object that the separation is bought by changing the physics.
We therefore add a reading on which nothing changes but the observation window.
The detection design is partitioned into four disjoint quarters of $[0,1]$, each with its $480$ points.
The scalar operator is fitted on each quarter separately, with its share of the same noise realization (Fig.~\ref{fig:window}).
Under a correct operator the pseudo-true value equals the true rate on every quarter, by an algebraic identity.
On the modal basis the first-order condition of \eqref{eq:nls} reads $\sum_n c_n^2k_n^2\,[G_n(\alpha)-H_n(\alpha)]=0$.
Here $G_n=\int t\,e^{-(\bar\alpha_n+\alpha)k_n^2t}\,\mathrm d\mu_t$ and $H_n=\int t\,e^{-2\alpha k_n^2t}\,\mathrm d\mu_t$.
When all $\bar\alpha_n$ coincide, every bracket vanishes at $\alpha=\bar\alpha$ before any integration in time, for every temporal design.
The flat line of Fig.~\ref{fig:window}(a) is thus a prediction with no free parameter, not a measurement.
Under three aggregated rates the pseudo-true value drifts by $27\%$ of $\alpha_0$, from $0.00931$ (first quarter) to $0.00664$ (last).
Over $30$ seeds, the per-window medians reproduce the analytic values to within $0.25$ standard errors.
To decide, we read $Q$ (Section~\ref{sec:qdef}) on the four estimates.
It rejects in every seed under single and under double misspecification.
Under correct specification it rejects in $4$ of $30$ seeds, a pre-registered negative against the ceiling.
On the pre-declared extension it rejects in $11$ of $200$ ($0.055$, $[0.028,0.096]$; Section~S9).
The modal head, correct on the same three-rate data, is invariant ($1/30$), so the drift follows the operator, not the data.
Fig.~\ref{fig:window}(b) maps the identified rate over sliding windows on $[0,2]$.
At the generating rates of the field, an evaluation that uses the truth, $T_5$ falls below $1$ past $t_0\approx1.25$.
Indeed, the fifth mode has decayed and the window alone removes it.
At the rate the single fit postulates on each window ($0.0066$ to $0.0093$), $T_5$ stays above $286$ on every cell.
A mode postulated at $0.007$ has not decayed where the true mode at $0.020$ has.
The last branch of Algorithm~\ref{alg:proc}, $\IN$ mute with $T_5\ge1$ and $Q$ firing, is exercised at the notch of Section~\ref{sec:notch}, where $Q$ rejects every seed and $\IN$ almost none.

\textbf{Two bounds on the separation.}
Both statistics are instruments for regular models, whereas neural networks are singular \cite{wei2023}.
The objection holds in weight space, not for \eqref{eq:imparts}--\eqref{eq:rank}, which are evaluated in the estimand (Section~\ref{sec:instr}).
For the second statistic, regularity is measured rather than assumed.
Indeed, $T_5<1$ states precisely that $I_{55}$ is too small to be treated as positive definite.
The non-identifiability induced here is structural (a vanishing column at $c_5=0$).
Practical non-identifiability along a curved valley~\cite{wieland2021} is not addressed.
Second, on detection alone cheaper statistics also succeed.
Three residual whiteness statistics, each with its own bootstrap null, need one pass where $\IN$ needs a $2000$-draw bootstrap.
They reject $1.00$ of the seeds under single misspecification and $0.067$ on the three correct designs.
We claim no advantage over them.
At the notch of Section~\ref{sec:notch}, they do better.
What none of them provides is the second reading, namely whether the coefficient was reachable.
\section{The Third Cause: Incomplete Optimization}\label{sec:networks}
The readings above come from least-squares surrogates, which have no optimization error and so isolate the operator and the design.
The networks below test whether the readings survive training.
This section reads three networks: a PINN frozen at its minimum, the same PINN trained on each window, and a network with one field per source.

\subsection{A network frozen at its minimum}\label{sec:frozen}
(H2) of Section~\ref{sec:h2} is an operating condition on the \emph{deployed} field.
We thus repeated the three cells with a trained network.
It is the PINN of Section~\ref{sec:estim}, fitted on the detection grid and frozen before any reading, as \cite{zhang2026} freeze theirs.
The residual is the network's, $r_i=y_i-u_\theta(x_i,t_i)$, whereas $\chi$ and $\partial^2_\alpha u$ remain those of the postulated ansatz.
The null is resampled around the deployed field with $\hat\alpha$ held fixed, as in the surrogate cells.
Under a correct operator the median is $0.220$.
The rejection rate is $0.067$ in sample, below the ceiling ($0.133$ out of sample).
Under single and double misspecification the medians are $130.6$ and $39.3$ against thresholds $2.67$ and $2.58$.
The frozen network is rejected in $30$ of $30$ seeds and $\hat\alpha$ settles at $0.0070$ and $0.0077$ (Table~\ref{tab:white}).
Specificity and power are thus both measured on a trained network.

\subsection{A network trained on each window}\label{sec:netwin}
Under three aggregated rates the analytic estimand drifts with the window (Section~\ref{sec:discrim}).
The network of Table~\ref{tab:white} was trained on each quarter with its data channel restricted to the window.
Its residual and initial-condition channels covered the whole domain.
It drifts with the window as the analytic head does.
Its median rate falls from $0.00902$ (first quarter) to $0.00651$ (last), within $3.1\%$ of the analytic pseudo-true values.
This fall, of the same sign, is $25\%$ of $\alpha_0$ and $15$ standard errors.
The whole-domain channels do not damp the drift.

\subsection{The hypothesis behind the specificity guarantee}\label{sec:h2}
The specificity of the statistic is the vanishing of its expected discrepancy $\mathbb E[d]$ under a correct operator only.
Write the residual $r=\varepsilon+\delta$, with $\delta$ its structural part.
Every term of $\mathbb E[d]$ carries $\delta$, and the plug-in variance absorbs its level, so the reading needs no reference.
$\mathbb E[d]$ vanishes with $\delta$, which requires a correct operator and a fit at its minimum.
The statistic tests the first; the second is what a trained network can miss.
We use the guarantee under a working hypothesis (H2): the state field does not respond to the rate, $\partial_\alpha u_\theta\equiv0$.
(H2) is declared, not proved, because it bounds the claim.
It holds by construction for the analytic-field and least-squares surrogates.
A trained network, whose free state absorbs part of the misfit, violates it.
Freezing the state at its minimum before testing (fit, freeze, test) restores it, and this is measured on an actual network (Section~\ref{sec:frozen}, Table~\ref{tab:white}, network rows).
It does not establish specificity \emph{through} training.
\begin{figure*}[t]\centering
\includegraphics[width=\textwidth]{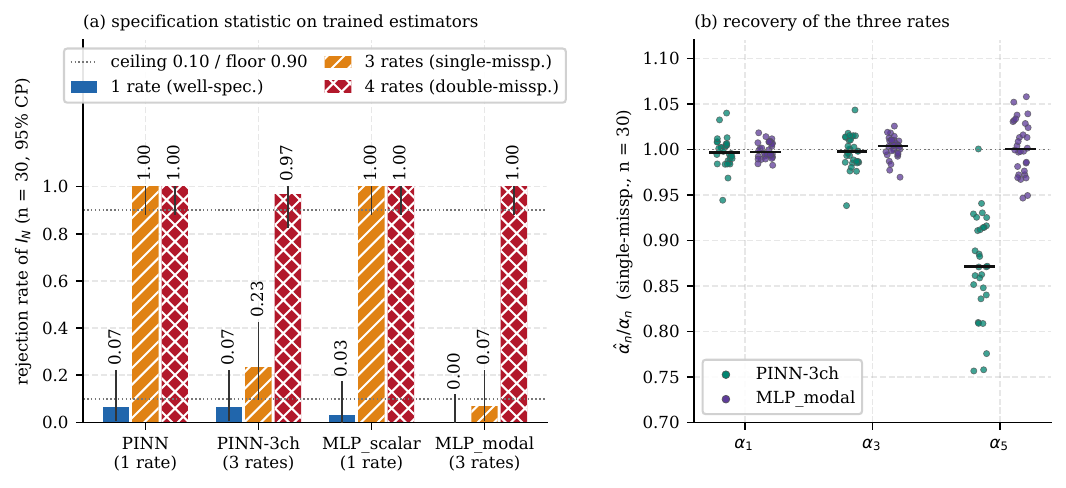}
\caption{Trained estimators, sources aggregated. (a) Rejection rate of $\IN$ ($n=30$, Clopper--Pearson $95\%$) per
estimator and generating configuration (PINN/well-specified from Table~\ref{tab:white}). (b) $\hat\alpha_n/\alpha_n$ per seed under
three rates: the three-channel PINN under-estimates $\alpha_5$; MLP\_modal recovers all three.}
\label{fig:networks}
\end{figure*}

The converse was measured too.
A three-channel network has one field per source with its own rate, and the sensor observes their sum.
It was trained under the same contract on the three-rate data, where its operator is correct.
Its frozen residual is rejected in $7$ of $30$ seeds ($0.233$, $[0.099,0.423]$; Fig.~\ref{fig:networks}).
On the pre-registered extension it is rejected in $30$ of $200$ ($0.150$, $[0.104,0.207]$).
Both are negatives against the ceiling.
The rejections follow the error on the fast, weak rate $\alpha_5$, under-estimated by $13\%$ ($4.5$ standard errors) in $97\%$ of the seeds.
On the same data, MLP\_modal, Hyb\_modal with its rates emitted by an $83$-parameter perceptron, recovers all three rates within one standard error.
Three
pre-registered training changes (fourfold L-BFGS budget, nominal initialization, per-channel residual weighting) leave this error at $8.9\%$, $10.1\%$ and $23.9\%$.
The fit stops short of its minimum, and its deterministic residual enters the statistic exactly as a wrong operator would.

On a network a rejection therefore reads \emph{wrong operator or unreached minimum}.
The network branches of Algorithm~\ref{alg:proc} separate the two by a second fit, of the analytic head on the same data.
That check lies outside the single-fit rule; it needs no retraining of the network and no oracle.
A Crank--Nicolson solver of the postulated operator serves in place of the closed form.
Its verdicts equal those of the closed form in all $180$ pre-registered readings ($30$ seeds, three configurations, in and out of sample).
On the same $200$ seeds MLP\_modal is rejected in $4$ ($0.020$, $[0.005,0.050]$), never where the network is not.
The network is rejected in $30$ ($0.150$; Fisher exact, one-sided as pre-registered, $p=1.3\times10^{-6}$).
A rejection the head does not share is optimisation, not operator, except at the notch of Section~\ref{sec:notch}, where the head is silent under a wrong operator too.
The same fork applies to the third reading.
On the network trained per quarter under a correct operator, the heterogeneity statistic rejects in $8$ of $30$ seeds at the surrogate's threshold.
On a pre-registered extension it rejects in $35$ of $200$ ($0.175$, $[0.125,0.235]$; supplementary material).
On a network it therefore says \emph{window-dependent estimand or window-dependent optimisation}.
The analytic head (scalar, least squares), without optimisation error, separates the two.
On the same $200$ seeds it is rejected in $11$, only $2$ of them where the network is not (Fisher exact, one-sided as pre-registered, $p=1.2\times10^{-4}$).
\section{Limitations and Negative Results}\label{sec:limits}

\subsection{A notch of the specification statistic}\label{sec:notch}
The power of $\IN$ is not monotone in the misspecification amplitude.
Since $\alpha_n=\alpha_0+h/k_n^2$ multiplies every mode by $e^{-ht}$, the family is a missing reaction term $-hu$.
The detection window $h\in[0.05,0.16]$ is a missing term of $0.5$ to $1.6$ times the dominant diffusive rate, $\alpha_0k_1^2=0.099$.
On this reaction--diffusion reference, in the deployed convention, the in-sample detection rate is $0.933$ at $h=0.160$.
It is $0.067$ ($[0.008,0.221]$) at $h=0.180$ and $0.367$ at $h=0.200$.
Out of sample: $1.000$, $0.133$ $[0.038,0.307]$, $0.167$.
A pre-registered extension of the grid restores full power at $h=0.230$, $0.260$ and $0.300$: $30$ of $30$ seeds, in and out of sample.
The analytic reading evaluates the statistic at the closed-form $\astar(h)$ instead of the fitted rate, with the same plug-in variance.
Its rate at $h=0.180$ is $0.000$: the notch does not come from fitting the rate.
The sixteen-point curve is in the supplementary material.
At a notch the expected indicator $\mathbb E[d]$ (Section~\ref{sec:h2}) vanishes under a wrong operator; the same test has a blind case in a linear model \cite[Section~II-C]{krauz2022}.
Why it vanishes at $h=0.180$ is left to future work.
On the same fits, the residual whiteness statistics reject $30$ of $30$ seeds at $h=0.180$.
So does the component of the test in the third power of the residuals.
Under a correct operator they reject $2$ to $3$ of $30$ (Table~S15).
Read jointly, $\IN$ and $T_5$ flag nothing there.
$\IN$ is mute, and nothing in the design is blind: its amplitudes are those of the detection design ($T_5$ was not evaluated along the sweep).
The third reading does.
The rate fitted on each quarter climbs from $0.0138$ to $0.0229$ (medians; $0.0189$ on the full window).
$Q$ rejects $30$ of $30$ seeds at $h=0.180$ and $0.200$, in and out of sample (pre-registered; Table~S15).

\subsection{Where the readings are mute on the window axis}\label{sec:windows}
Three further limits were pre-registered on the window axis, and one rule produces all three.
A missing dynamic is visible to $\IN$ only while its residual has had time to accumulate and before it has decayed.
At the origin of a window the structural residual and the sensitivity both vanish linearly in $t$.
Thus the signal available to the statistic grows as $t^3$.
At long times a fast mode has died.
Consequently, (i) the first quarter is the hardest cell for a slow missing structure.
On $[0,\tfrac14]$ the power of $\IN$ is $0.87$ under single and $0.67$ under double misspecification ($0.92$ and $0.66$ over $200$ seeds).
The heterogeneity statistic on the whole partition is $1.00$.
(ii) A missing fast mode is seen on the first quarter only (Section~\ref{sec:return}).
(iii) The heterogeneity statistic is blind to a missing mode orthogonal to the postulated basis on the design.
Indeed, the normal equations of the modal head do not see it.
This was predicted before the run and confirmed ($1/30$, $5/200$), and the two readings are complementary.
On the equispaced grid the information matrix of the rates is diagonal, so $T_5$ is a closed function of $c_5$ and $\hat\alpha$.
The second way to be blind, rates not separable on the design, is not exercised here (supplement, Section~S12).
The identifiability reading is local on this axis.
Non-identifiability can come from the decay of the mode over the window rather than from its amplitude.
At the postulated rate, the reading then does not see it (no cell below $1$ against four at the generating rates).
The variant at the fitted rates of the alternative, $T_5(\hat\theta)$, wanders with $\hat\alpha_5$.
It declares $40$ of $90$ non-identifiable pairs identifiable.
We do not claim here the separation of scales of Section~\ref{sec:discrim}.

\subsection{Other pre-registered negatives, and scope}\label{sec:scope}
Table~\ref{tab:negatives} collects the pre-registered claims that fell on the wrong side of their threshold.
The failed $\chi^2(1)$ calibration, whose criterion was written with its run, is reported in Section~\ref{sec:zero}.
Each is reported as a negative and none is requalified; the recovery negative is against explicit regularization \cite{tikhonov1995}.
All results rest on one self-adjoint parabolic operator with absolute Gaussian noise, one detection design, its time windows and six noise levels.
$\IN$, the heteroscedasticity component of the test (Section~\ref{sec:instr}), would also fire under a correct operator on noise whose variance covaries with $\chi^2$, or on correlated noise.
A rejection then concerns the noise model as much as the operator.
Under AR(1) noise in time the i.i.d.\ bootstrap null over-rejects ($6$ and $11$ of $30$ at $\rho=0.5$ and $0.8$).
A block bootstrap keeps at most $3$ of $30$ and full power (Table~S13).
Heteroscedastic noise was not run.

\begin{table}[t]
\centering
\caption{Pre-registered claims rejected by their own threshold; $n=30$ unless stated.}
\label{tab:negatives}
\footnotesize\setlength{\tabcolsep}{3pt}
\begin{tabular}{p{0.36\columnwidth}p{0.2\columnwidth}p{0.36\columnwidth}}
\toprule
claim & threshold & measured \\
\midrule
network beats oracle-tuned Tikhonov at field recovery & error ratio $\le0.70$ & $4.03$, $2.45$, $1.79$ \emph{worse} at $\sigma=0.01,0.02,0.05$ \\
field head collapses into a flat anisotropic valley (S6) & $\kappa\ge10$ & $\kappa=2.27$ ($2.08$--$2.69$; $2.57$ under correct specification; floor $2$) \\
modal field is ``free'' flexibility & fits no better than constants & fits better \\
three-channel network, correct operator: $\IN$ specific & rejection $\le0.10$ & $7/30$, $30/200$; $5/30$, $6/30$, $22/30$ after training changes \\
per-quarter network, correct operator: $Q$ specific (surrogate's threshold) & rejection $\le0.10$ & $8/30$; $35/200$ \\
surrogate, correct operator: $Q$ specific & rejection $\le0.10$ & $4/30$; $11/200$ (extension) \\
$\IN$ power along the $h$ sweep, once detected & detection $\ge0.90$ & $0.067$ at $h=0.180$, $0.367$ at $h=0.200$ \\
two closed-form predictions on the notch (location; local power of $\IN$ along $h$) & pre-registered bands & both outside: $h_c=0.208$; power $0.967$ at $h=0.075$. Their analysis is outside this paper \\
fourth cell, $c_5=0$: $\IN$ fires & $\ge0.90$ & $19/30$; $T_5=0$ as predicted \\
\bottomrule
\end{tabular}
\end{table}

\section{Reproducibility and Provenance}\label{sec:prov}
\paragraph{Code and data availability} The claim-to-artefact map (\claimmaptab{} of
the supplementary material), the certified artefacts and the dated pre-registrations are deposited at \OSFANON.
The analysis and certification code \ZENODO.
Four limitations of the record are declared, each with its commits in Section~S16 of the supplementary material.
No threshold was adjusted after seeing a result.
\section{Conclusions: Diagnose Before Interpreting}\label{sec:conclusion}
On an exactly solvable benchmark the coefficient settles $30\%$ off while the in-domain error is $1.36$ times the noise, and three questions put to the fitted model say what to change: the model, the experiment, or the training.
Box described scientific learning as an alternation of estimation and criticism \cite{box1980}; physics-informed learning has automated the first, and the diagnostic supplies the second.
The model is fitted, challenged, and only then revised or used for interpretation.

\textbf{What the diagnostic establishes.}
On this benchmark a small prediction error does not validate a learned coefficient.
Three causes of a wrong one are told apart without the truth: a wrong operator, a design blind to a named alternative, and a network short of its minimum.

\textbf{What it needs.}
It needs a postulated operator whose sensitivity in the coefficient can be computed in closed form, from a solver or by automatic differentiation.
The design reading also needs a richer operator that the practitioner names and never fits.
A network must be frozen at its minimum, or else a second fit of the analytic head is made on the same data.
It needs no truth, and only the window reading refits.

\textbf{What it does not claim.}
Its silence is not a proof of validity: it says what this design could not see.
Its limits are those of Section~\ref{sec:limits}.
Power falls in a notch, reached at $h=0.180$ and left by $h=0.230$, and the window reading sees through it.
Power is lost beyond $7\%$ noise.
The null is a block null under correlated noise, and the rank statistic is local on the window axis.
The benchmark is one exactly solvable operator; porous walls~\cite{luikov1975} and parallel thermal paths make its case ordinary.

A learned coefficient is a hypothesis about the physics, not a measurement of it.
Challenging it before reading it should be as routine as validating a prediction before using it.
\section*{Acknowledgment}
The line of inquiry and the question this paper asks are the author's.
So is its articulation, from the success of the field through the silent hypothesis, the diagnostic and its controls to the three causes and the scope.
So are the exactly solvable operator and the two axes on which it is varied.
The demonstration follows the author's order: instrument, zero, plateau, return of the zero, procedure.
Every pre-registered threshold and decision rule is the author's, as is every arbitration.
So is the reading of each result, the negatives reported here included.
Large language models \cite{claudeai,mistralai} assisted in drafting the manuscript, every section and the reference list included, and its code.
These were Claude (Anthropic: Opus, Sonnet and Fable families) and Mistral Vibe (Mistral AI), in 2025--2026.
They were used for writing and implementation, not as authors.
The author verified each reference against its primary source and each number against its artefact, at the tier the claim-to-artefact map gives it.
The author takes full responsibility for the content.

\end{document}